\documentclass[runningheads]{llncs}
\usepackage{graphicx}
\usepackage{multirow}
\usepackage{amssymb}
\usepackage{threeparttable}

\begin{document}
%

\title{Towards an efficient framework for Data Extraction from Chart Images}

\titlerunning{Towards an efficient framework for Data Extraction from Chart Images}

\author{Weihong Ma\inst{1} \and Hesuo Zhang\inst{1} \and 
Shuang Yan\inst{2} \and Guangshun Yao\inst{2} \and
Yichao Huang\inst{2} \and Hui Li\inst{3}      \and Yaqiang Wu\inst{3} 
\and Lianwen Jin\inst{1,4}
}

\authorrunning{ et al.}
%

\institute{
South China University of Technology, Guangzhou, China \\
\email{\{eeweihong\_ma, eehesuo.zhang\}@mail.scut.edu.cn, eelwjin@scut.edu.cn} \and
IntSig Information Co.,Ltd., Shanghai, China \\
\email{\{shuang\_yan, guangshun\_yao, charlie\_huang\}@intsig.net} \and
Lenovo Research, Beijing, China\\
\email{\{lihuid, wuyqe\}@lenovo.com} \and
Guangdong Artificial Intelligence and Digital Economy Laboratory (Pazhou Lab), Guangzhou, China
}

%
\maketitle              
\begin{abstract}

In this paper, we fill the research gap by adopting state-of-the-art computer vision techniques for the data extraction stage in a data mining system.
As shown in Fig.~\ref{fig1}, this stage contains two subtasks, namely, plot element detection and data conversion.
For building a robust box detector, we comprehensively compare different deep learning-based methods and find a suitable method to detect box with high precision.
For building a robust point detector, a fully convolutional network with feature fusion module is adopted, which can distinguish close points compared to traditional methods.
The proposed system can effectively handle various chart data without making heuristic assumptions.
For data conversion, we translate the detected element into data with semantic value.
A network is proposed to measure feature similarities between legends and detected elements in the legend matching phase.
Furthermore, we provide a baseline on the competition of Harvesting raw tables from Infographics.
Some key factors have been found to improve the performance of each stage.
Experimental results demonstrate the effectiveness of the proposed system.

\end{abstract}

\section{Introduction}

\begin{figure}
    \centering
    \includegraphics[width=\textwidth]{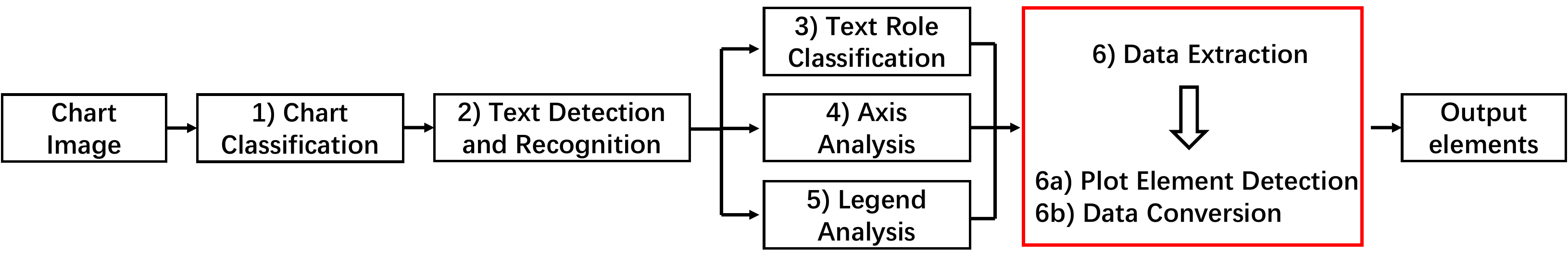}
    \caption{Generic pipeline for extracting data in a data mining system. We mainly discuss the sixth stage (data extraction) assuming that the previous output has been obtained.} \label{fig1}
\end{figure}

Chart data is one of the important information transmitted medium that clarifies and integrates difficult information concisely~\cite{siricharoen2013infographics}.
In recent years, an increasing number of chart images have emerged in multimedia, scientific papers, and business reports.
Therefore, the issue of automatic data extraction from chart images has gathered significant research attention~\cite{boschen2017comparison,liu2013review,mei2018design,purchase2014twelve}.

\begin{figure}
    \centering
    \includegraphics[width=\textwidth]{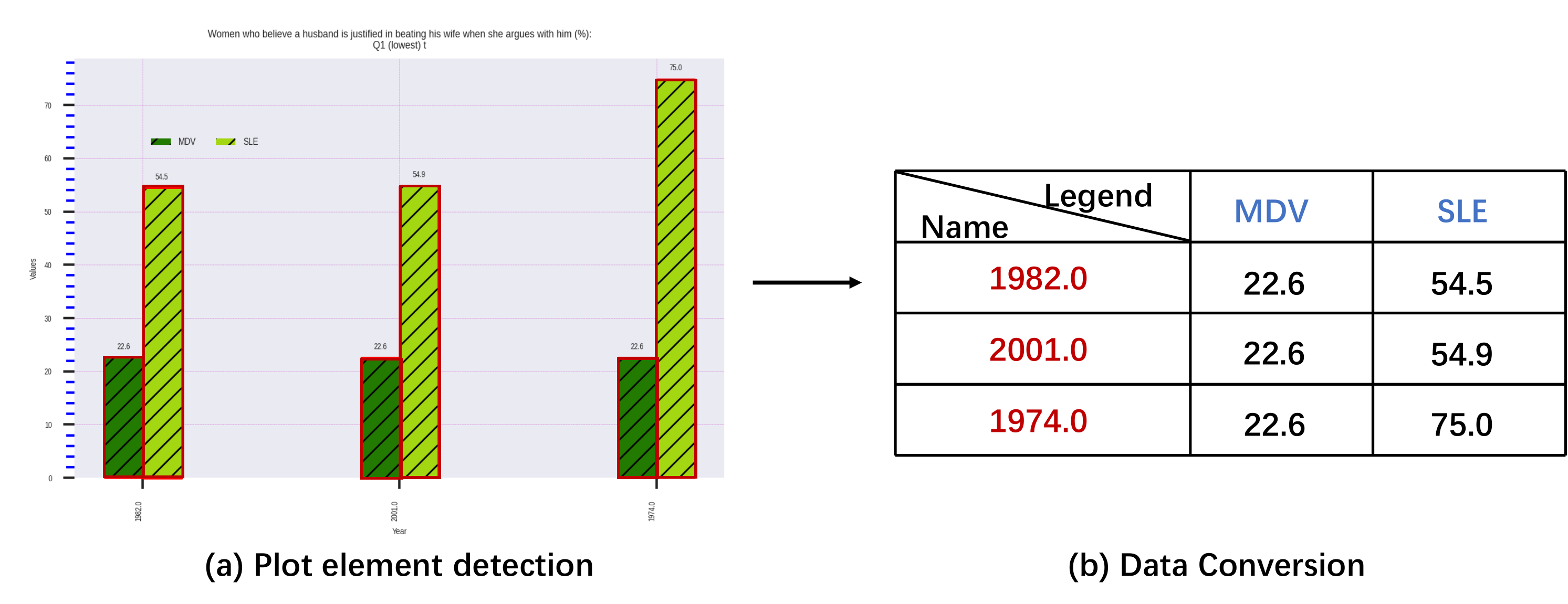}
    \caption{Illustration of the data extraction stage.} \label{fig1_1}
\end{figure}

As shown in Fig.~\ref{fig1}, in general, a chart data mining system~\cite{davila2019icdar} includes the following six stages: chart classification, text detection and recognition, text role classification, axis analysis, legend analysis and data extraction.
Among all the aforementioned stages, data extraction is conducted as the most crucial and difficult part, whose performance depends on the quality of localization.
In this work, we mainly discuss the data extraction stage.
The goal in this stage is to detect elements in the plot area and convert them into data marks with semantic value.
As shown in Fig.~\ref{fig1_1}, this task has two subtasks: plot element detection and data conversion.

\begin{figure}
    \centering
    \includegraphics[width=0.8\textwidth]{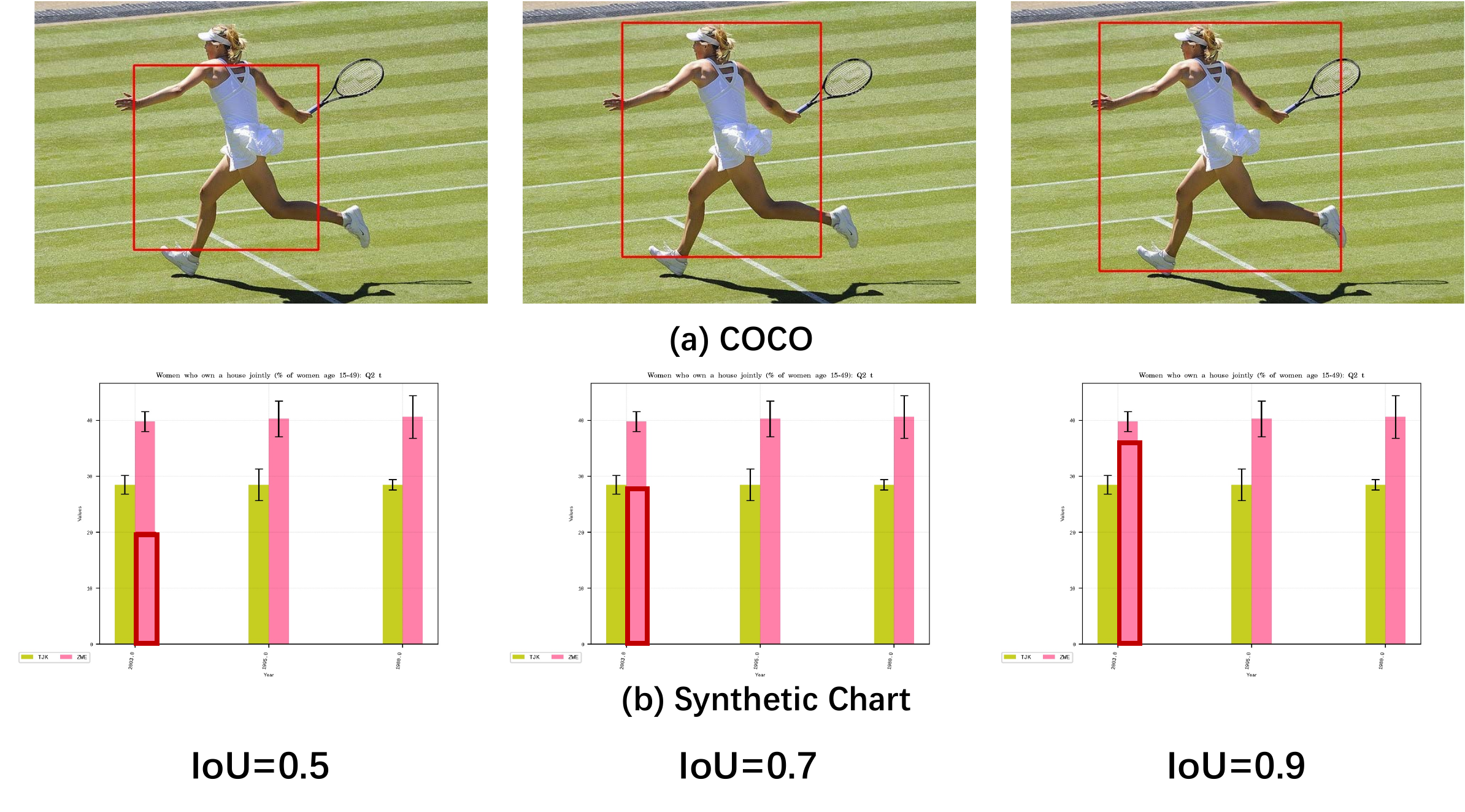}
    \caption{Visualization when intersection-over-union (IoU) values range from 0.5 to 0.9 on (a) COCO and (b) synthetic chart images. An IoU value of 0.5 is acceptable on natural images; a higher IoU value, such as 0.7, is redundant. However, for chart images, such values are unacceptable. Even if the IoU is 0.9, there is still a small numerical deviation on bar charts.}\label{fig2}
\end{figure}

To build a robust data extraction system, we can learn methods from the field of object detection.
However, it should be clear that chart images differ significantly from natural images.
As shown in Fig.~\ref{fig2}, (a) is an image from COCO dataset~\cite{lin2014microsoft}, and (b) is an image from synthetic chart dataset~\cite{davila2019icdar}.
First, compared with general object, elements in the chart images have a large range of aspect ratios and sizes.
Chart images contain a combination of different elements.
These elements can be either very short, such as numerical tick points or long, such as in-plot titles.
Second, chart images are highly sensitive to localization accuracy.
While the intersection-over-union (IoU) values in the range of 0.5 to 0.7 are acceptable for general object detection, it is unacceptable for chart images.
As shown in Fig.~\ref{fig2}b, even when the IoU is 0.9, there is still a small numerical deviation on bar images, which shows the sensitivity of chart images to IoU.
Therefore, for chart data extraction, highly precise bounding boxes or points, i.e., with high IoU values are required for the detection system.

Currently, state-of-the-art computer vision techniques have not been fully adopted by chart mining approaches.
Moreover, there have been very few comparisons using deep learning-based methods for chart mining.
It is believed that deep learning-based methods can avoid hard heuristic assumptions and more robust when handling various real chart data.
In this study, using published real-world datasets$\footnote[1]{http://tc11.cvc.uab.es/datasets/ICPR2020-CHART-Info\_1}$, we attempt to fill this research gap in the data extraction stage. 
In the proposed framework, elements in the main plot area are first detected.
Based on axis analysis and legend analysis results from previous stages in a data mining system,
we then convert detected elements into data marks with semantic value.
The contribution of this work can be summarized as follows.
(\romannumeral1) For building a robust box detector, we comprehensively compare different deep learning-based methods.
We mainly study whether existing object detection methods are adequate for box-type element detection.
In particular, they should be capable of (a) detecting elements with a large aspect ratio range and (b) localizing objects with a high IoU value.
(\romannumeral2) For building a robust point detector, we use a fully convolutional network (FCN) with feature fusion module to output a heatmap mask.
It can distinguish close points well while traditional methods and detection-based methods easily fail.
(\romannumeral3) For data conversion, 
in the legend matching phase, a network is trained to measure feature similarities.
It is robust than image-based features when noise exists in feature extracting phase.
Finally, we provide a baseline on a public dataset which can facilitate further research.
Experimental results demonstrate the effectiveness of the proposed system.
The implementation of our pipeline will be available to the public for reproducing these results.

\section{Related work}
In this section, we review previous works on data extraction in a chart mining system.
We mainly focus on related works of classification, element detection and data conversion.

According to the types of detection data, we can divide the chart data into box-type and point-type data.
Box-type data includes bars and boxplots.
These charts are commonly used to visualize data series that have a categorical independent variable.
For the task of box detection, 
some methods have been proposed to detect elements through the characteristics of bars~\cite{al2017machine,balaji2018chart,dai2018chart,savva2011revision}.
Assuming that the bar mark is solidly shaded using a single color, Savva et al.~\cite{savva2011revision} used connected component (CC) analysis method and heuristic rules to extract the data mark.
Balaji et al.~\cite{balaji2018chart} and Dai et al.~\cite{dai2018chart} also used image processing methods to detect bars.
They first obtained the binary image and used open morphological operation to filter noise.
Next, they performed the CC labeling algorithm to find the bars.
Rabah et al.~\cite{al2017machine} used heuristic features based on shape, pixel densities, color uniformity, and relative distances to the axes.
However, these methods may fail when detecting small bars.
In response to this problem,
some methods based on deep neural networks use object detection architectures to locate the bars~\cite{choi2019visualizing,liu2019data,Methani2020Data}, which are more robust to extract features.

Point-type data, including charts such as scatter, line, and area, are semantically similar because they present one or more sequences of 2D points on a given Cartesian plane.
The scatter chart is the most basic type in these charts.
A line chart is created when the points are connected to form curves.
An area chart highlights the area under the curve, typically using coloring or texture patterns.
Assuming that only detected data and text elements are in the plot area,
Khademul et al.~\cite{molla2003line} proposed a system for extracting data from a line chart.
The system first detected axes by projection method and then used CC analysis to filter text elements inside the plot area.
Finally, data are extracted using a sequential scanning process.
Considering that the chart might have grid lines, Viswanath et al.~\cite{reddy2015image} proposed an image processing-based method and developed a semi-supervised system.
However, the method assumes that grid lines should not be more visually distinctive than the data marks.
Thus, these approaches make multiple assumptions and often fail when processing images with a complex background.
Cliche et al.~\cite{cliche2017scatteract} used object detection model~\cite{stewart2016end} to detect scatter points.
This method is more robust than the image processing-based method.
However, this method may fail when points are close to each other.

In the data conversion stage, legend analysis and axis analysis should be obtained. 
If legends exist, the shape and color of elements will be used to identify data marks by data series.
Choudhury et al.~\cite{ray2016curve} proposed clustering based on shape and color for line graphs.
Each cluster is output as a curve.
Using the axes analysis results, these relative coordinates can be projected onto the original data space.
To recover data from the bar chart, Savva et al.~\cite{savva2011revision} considered linear mapping and calculated the scaling factor between image space and data space.

\begin{figure}
    \centering
    \includegraphics[width=\textwidth]{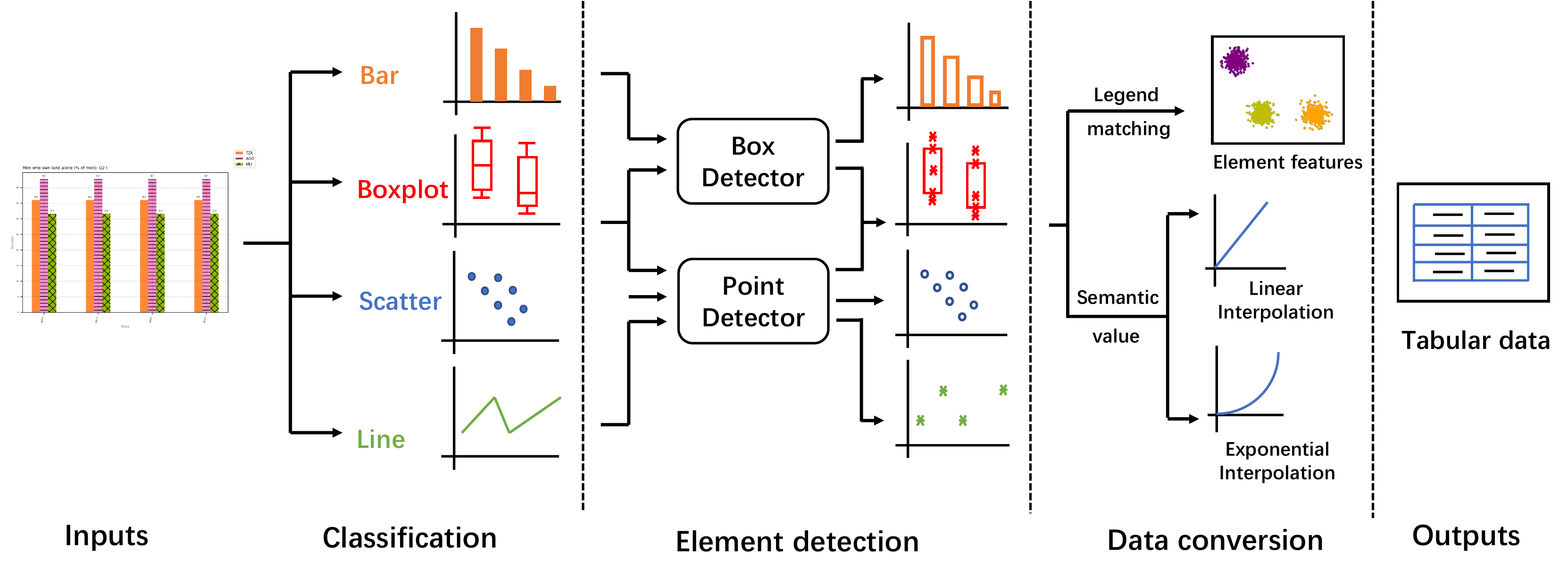}
    \caption{The overall architecture of the proposed framework. First, the input chart image is classified by the pre-trained classification model. Second, two detectors, named box detector and point detector, are built for different chart types. Third, in legend matching phase, elements are divided into corresponding legends by comparing their features similarities. Their semantic value are calculated by interpolation method. Finally, the value of elements are output into tables.}\label{fig3}
\end{figure}

\section{Methodology}
The overall architecture of our proposed method is presented in Fig.~\ref{fig3}.
Functionally, the framework consists of three components: a pre-trained chart classification model, element detection module for detecting box or point, and data conversion for determining element values.
In the following sections, we first introduce the details of the box and point detectors.
Next, we provide implementation details of the data conversion.

\subsection{Box detector}
\begin{figure}
    \centering
    \includegraphics[width=\textwidth]{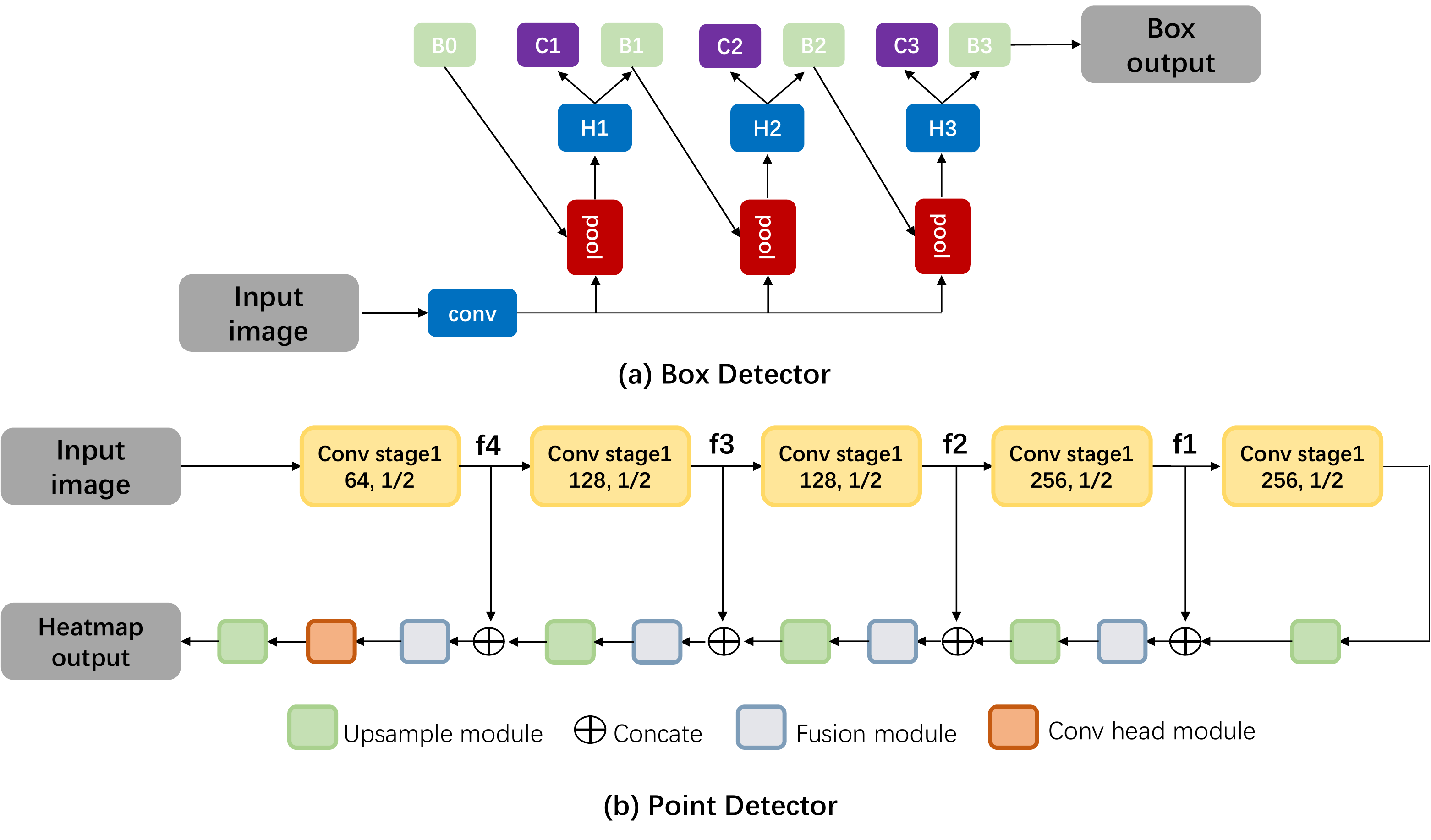}
    \caption{Network architecture of box detector and point detector.} \label{fig4}
\end{figure}

To extract robust features at different scales, we use a ResNet-50~\cite{he2016deep} with a feature pyramid network (FPN)~\cite{lin2017feature}.
FPN uses a top-down architecture with lateral connections to fuse features of different resolutions from a single scale input, enabling it to detect elements with a large aspect ratio range.
To detect a box with high IoU, we choose the Cascade R-CNN~\cite{cai2018cascade} as our box detector.
As shown in Fig.~\ref{fig4}(a), the box detector has four stages, one region proposal network (RPN) and three for detection with IoU={0.5, 0.6, 0.7}.
The sampling of the first detection stage follows~\cite{ren2016faster}.
In the following stages, resampling is implemented by simply using the regressed outputs from the previous stage.

\subsection{Point detector}
Points are another common chart elements in chart data.
As mentioned earlier, the corresponding chart types include scatter, line, and area.
Generally, points are densely distributed in the plot area, and the data is represented as the format of $(x, y)$.
In this work, we use segmentation-based method to detect points, which can help distinguish close points.

\subsubsection{Network architecture}
As shown in Fig.~\ref{fig4}(b), 
four levels of feature maps, denoted as $f_{i}$, are extracted from the backbone network, whose sizes are $1/16$, $1/8$, $1/4$ and $1/2$ of the input image, respectively.
Then, the features from different depths are fused in the up-sampling stage.
In each merging stage, the feature map from the last stage is first fed to the up-sampling module to double its size, and then concatenated with the current feature map.
Next, the fusion module, which is built using two consecutive $conv_{3\times3}$ layers, produces the final output of this merging stage.
Following the last merging stage, the head module, built by two $conv_{3\times3}$ layers, is then used.
Finally, the feature map is upsampled to the image size.

\subsubsection{Label Generation}
To train the FCN network, we generate a heatmap mask.
The binary map, which sets all pixels inside the contour to the same value, can not reflect the relationship between each pixel.
In contrast to the binary segmentation map, we draw Gaussian heatmap for these points on the mask.
The Gaussian value ${Y_{xy}}$ is calculated using the Gaussian kernel function.
If two Gaussians overlap and one point has two values, we use the maximum value.
\begin{equation}
    Y_{xy} = e^{- \frac{(x-p_{x})^{2}+(y-p_{y})^{2}}{2\sigma^{2}}}
\end{equation}
where ${(x,y)}$ is the point coordinate on the mask, $(p_{x}, p_{y})$ is the center of the target point.
${\sigma}$ is a Gaussian kernel parameter that determines the size.
Here, we set the value of ${\sigma}$ as 2.

\subsubsection{Post-processing}
In the testing phase, the point detector outputs a heatmap mask.
We first filter the output noise outside the main plot area.
Then, we use a high confidence threshold to output positive region.
The final points output are obtained by finding the center of connected components.
In the process of connected component analysis, for a larger connected area, we also randomly select points inside the area as the output.

\subsection{Data Conversion}
\begin{figure}
    \centering
    \includegraphics[width=\textwidth]{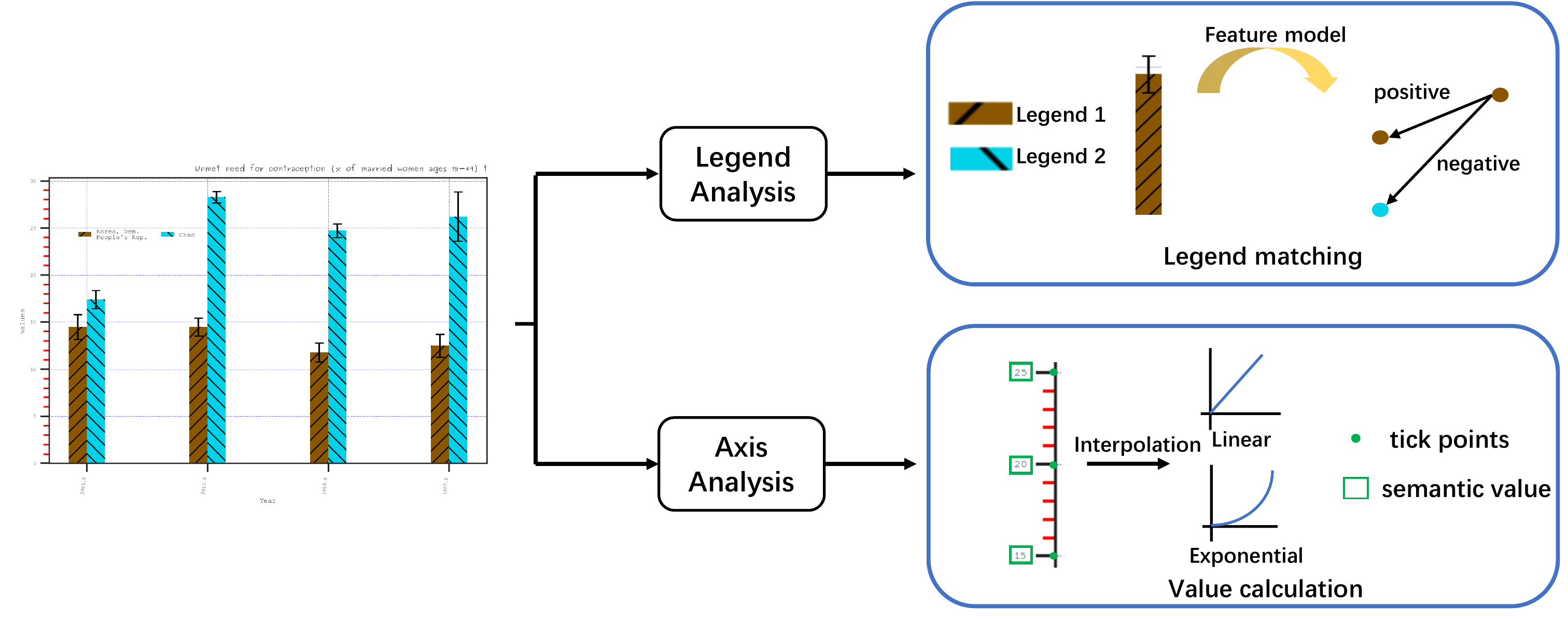}
    \caption{Pipeline of Data Conversion.} \label{fig5}
\end{figure}

After detecting the elements, we need to determine the element values.
In this stage, the goal is to convert detected elements in the plot area into data marks with semantic value.
As shown in Fig.~\ref{fig5}, legend matching and value calculation are performed in this stage.

\textbf{Legend matching: }
Based on the legend analysis result, which is obtained from the fifth stage in a data mining system, 
we can get the position of the legends.
If there exists legends, we need to extract features of elements and legends.
Then we use L2 distance to measure feature similarities and divide elements into corresponding legends.
Image-based features, such as RGB features and HSV features, is not robust when the detection result is not tight enough.
Therefore, we propose to train a feature model to measure feature similarity.

The network directly learns a mapping $f(x)$ from patch input image $x$ into embedding vectors $\mathbb{R}^{d}$.
It is comprised of multiple modules, which are built using conv-BN-ReLU layers, and finally outputs a 128-D embedding vector for each patch input.
In the training phase,
the network is optimized using a triplet-based loss~\cite{schroff2015facenet}.
This loss aims to separate the positive pair from the negative by a distance margin.
Embedding vectors of the same cluster should have small distances and different clusters should have large distances.
In the testing phase, the cropped legend patch and element patch are fed into the model.
For each element, the legend with the smallest distance on feature dimension is the corresponding class.

\textbf{Value calculation: }
Based on the axis analysis result, which is obtained from the forth stage,
we can get the position of the detected tick points and their corresponding semantic values.
Then, we analyze the numerical relationship between adjacent tick points, including the case of linear or exponential.
Finally, we calculate the value of the unit scale and use the interpolation method to determine the element value.

\section{Experiments}
\subsection{Datasets}

\begin{table}[h]
    \footnotesize
    \centering
    \caption{Distribution of the two datasets used for the experiments}
    \label{table:Dataset split}

    \begin{tabular}[!hbt]{|c|c|c|c|c|}
        \hline
        Dataset & Chart type & Num. of training set & Num. of validation set & Total Num.\\
        \hline

        \multirow{4}*{Synth2020} & Bar & 4,264 & 536     & 4,800 \\
        \cline{2-5}
                                 & Boxplot & 2,132 & 268 & 2,400 \\
        \cline{2-5}
                                 & Line  & 1,066 & 134   & 1,200 \\
        \cline{2-5}
                                 & Scatter & 1,066 & 134   & 1,200 \\
        \hline

        \multirow{4}*{UB PMC2020}  & Bar & 781 & 98  & 879 \\
        \cline{2-5}
                                 & Boxplot & 174 & 22 & 196 \\
        \cline{2-5}
                                 & Line & 658 & 83 & 741 \\
        \cline{2-5}
                                 & Scatter & 272 & 35  & 307 \\
        \hline
    \end{tabular}
\end{table}

There are two sets of datasets used in this work, which are named Synth2020 and UB PMC2020, respectively.
The earlier version of these two datasets are published in~\cite{davila2019icdar}.
Thanks to the ICPR2020 Competition on Harvesting Raw Tables from Infographics$\footnote[1]{https://chartinfo.github.io}$, 
more challenging images have been added and the size of the dataset has been extended.
The first dataset, Synth2020, is the extended version of Synth2019.
Multiple charts of different types are created using the Matplotlib library.
The second dataset is curated from real charts in scientific publications from PubMedCentral~\cite{davila2019icdar}, which has different image resolutions and more uncertainties in the images.
We divide the official training dataset into a training set and validation set with a ratio of 8:1 randomly.
Details of the split of these two datasets are given in Table \ref{table:Dataset split}.
The specific training and validation sets will be published.

\subsection{Evaluation}
We use the competition evaluation script$\footnote[2]{https://github.com/chartinfo/chartinfo.github.io/tree/master/metrics}$ to measure the model performance.
Two metrics for measuring the detector performance and data conversion performance are proposed.
For different types of chart data, the script has a different evaluation mechanism.
Details of the evaluation mechanism are mentioned in the competition.
For box detection evaluation, to accurately evaluate the model performance under different IoU, we also refer to the field of object detection and calculate F-measure when IoU is equal to 0.5, 0.7, and 0.9.

\subsection{Implementation details}
The synthetic and UB PMC datasets are trained and tested separately.
In this section, we introduce the details of our implementation.

In the box detector experiment, we choose bar type data for training.
The backbone feature extractor is ResNet-50 pre-trained on ImageNet.
In the regression stage, we adopt RoIAlign to sample proposals to a 7x7 fixed size.
The batch size is 8 and the initial learning rate is set to 0.01.
The model is optimized with stochastic gradient descent(SGD) and the maximum epochs for training is 20.
In the inference stage, non-maximum suppression (NMS) is utilized to suppress the redundant outputs.

In the point detector experiment, we choose scatter type data for training.
In the training stage, 
we use MSE loss to optimize the network.
Multiple data augmentations are adopted, including random crop, random rotate, random flip, and image distortion, to avoid overfitting.
We adopt the OHEM~\cite{shrivastava2016training} strategy to learn hard samples.
The ratio of positive and negative samples is 1:3.
The model is optimized with Adam optimizer and the maximum iterations is 30k with a batch size of 4.

In the data conversion experiment, we train the model to extract features for clustering.
The input size for training is 24x24, and the embedding dimension is set to 128.
The model is optimized with Adam optimizer and the maximum iterations is 50k.
The batch size is 8 and the initial learning rate is set to 0.001.

\subsection{Result and Analysis}

\subsubsection{Evaluation of Box detector}
In this section, the performance of the box detector is evaluated in terms of Score\_a and F-measure when the value of IoU is set to 0.5, 0.7, 0.9, respectively.
Score\_a uses the evaluation mechanism from ICPR2020 competition.
The trained models are tested on the Synth2020 validation set and UB PMC2020 testset, respectively.
Since the testset of Synth2020 is currently unavailable, we use validation set to test the model performance on the Synth2020 dataset.

\begin{table}[h]
    \footnotesize
    \centering
    \caption{Evaluation results of box detector on bar type data}
    \label{table:box detector}

    \begin{tabular}[!hbt]{|c|c|c|c|c|c|}
        \hline
        Dataset & Model & \  IoU=0.5 \ & \  IoU=0.7 \  & \  IoU=0.9 \  & \  Score\_a \  \\
        \hline

    \multirow{5}*{Synth2020 validation bar} & SSD & 80.98 & 69.56 & 28.79 & 67.87 \\
        \cline{2-6}
                                 & YOLO-v3 & 85.30 & 76.54 & 38.55 & 90.96 \\
        \cline{2-6}
                                 & Faster R-CNN & 94.80 & 92.47 & 48.92 & 92.33 \\
        \cline{2-6}
                                 & Faster R-CNN+FPN & 96.46 & 94.39 & 52.30 & 92.89 \\
        \cline{2-6}
                                 & Cascade R-CNN+FPN & \textbf{96.86} & \textbf{96.25} & \textbf{93.97} & \textbf{93.36} \\
        \hline

    \multirow{5}*{UB PMC2020 testset bar}  & SSD & 43.65 & 26.28 & 2.67 & 25.83 \\
        \cline{2-6}
                                 & YOLO-v3 & 58.84 & 36.14 & 4.14 & 60.97 \\
        \cline{2-6}
                                 & Faster R-CNN & 66.37 & 60.88 & 29.13 & 70.03 \\
        \cline{2-6}
                                 & Faster R-CNN+FPN & 85.81 & 78.05 & 31.30 & 89.65 \\
        \cline{2-6}
                                 & Cascade R-CNN+FPN & \textbf{86.92} & \textbf{83.53} & \textbf{55.32} & \textbf{91.76} \\
        \hline
    \end{tabular}
\end{table}

\begin{figure}
    \centering
    \includegraphics[width=0.93\textwidth]{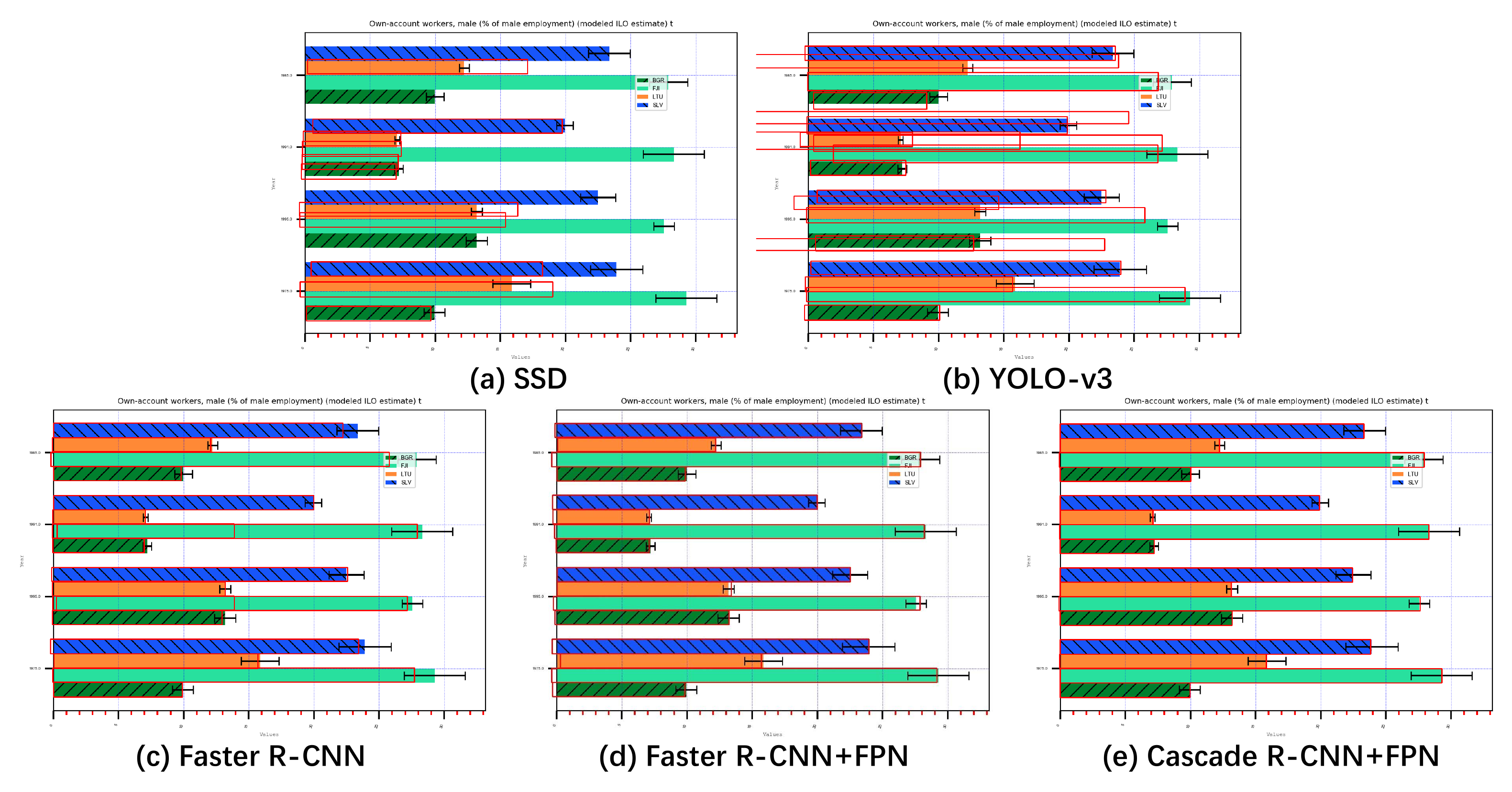}
    \caption{Detection result of different models on an example bar image from Synth2020.} \label{fig6}
\end{figure}

For comparison, we implement different detection models, including one-stage and two-stage detection models.
The one-stage models are SSD~\cite{liu2016ssd} and YOLO-v3~\cite{farhadi2018yolov3}, whereas the two-stage model is Faster R-CNN~\cite{ren2016faster}.
As listed in Table \ref{table:box detector},
the performance of the one-stage model performs worst, and the multi-stage regress heads help to obtain high accuracy.
Furthermore, the additional FPN structure effectively helps to detect elements with a large aspect ratio range.
On both Synth2020 and UB PMC2020 dataset, the Cascade R-CNN model with FPN structure performs the best.
Therefore, for bar type data detection, models with multiple regression heads and FPN structure achieve impressive performance.

One-stage models output poor results in earlier iterations.
At the same time, NMS can not filter these error outputs effectively, which can be best viewed in Fig.~\ref{fig6}(b).
NMS can not suppress these outputs because the IoU between these long rectangles is smaller than 0.5.
Owing to these reasons, the model can not reach the global optimal solution.

\subsubsection{Evaluation of Point detector}

\begin{table}[h]
    \footnotesize
    \centering
    \caption{Evaluation results of point detector on scatter type data}
    \label{table:point detector}
    \begin{tabular}{|c|c|c|c|c|c|}
        \hline
        & \  CC based \ & \  Detect based \ & \  Pose ResNet \ &  \ Proposed \ \\
        \hline
        Synth2020 validation set  & 57.84  & 72.34  & 78.98 & \textbf{87.20} \\
        \hline
        UB PMC2020 validation set & 53.36  & 82.12  & 82.46 & \textbf{86.46} \\
        \hline
        UB PMC2020 testset  & 51.58  & 84.54  & 84.58 & \textbf{88.58} \\
        \hline
    \end{tabular}
\end{table}

\begin{figure}
    \centering
    \includegraphics[width=\textwidth]{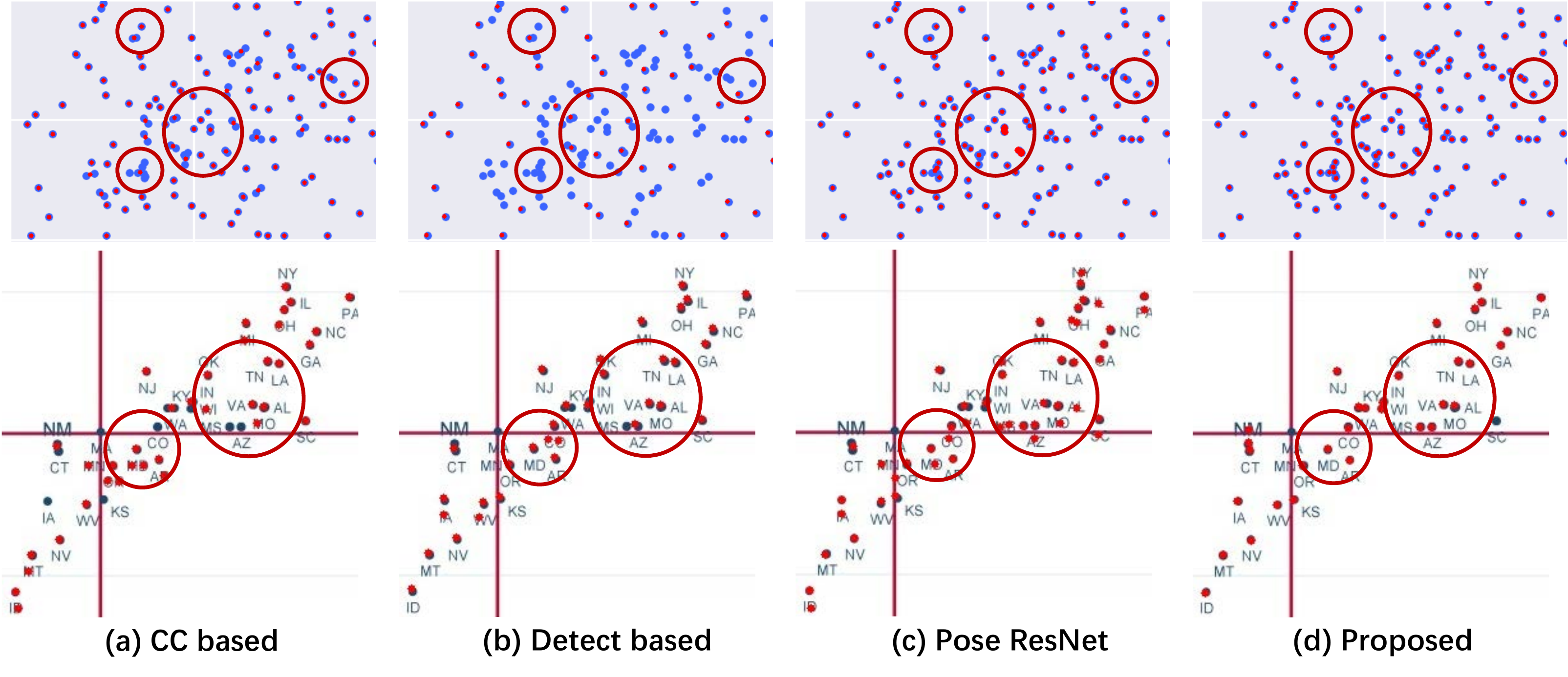}
    \caption{Visualization of the detection results on chart data from different models. Patches are shown for visualization. The detected points are drawn as red dots on the input image. The data of first row is from Synth2020 and the data of second row is from UB PMC2020. Circles in red show some key differences between these models.} \label{fig7}
\end{figure}

In this section, the performance of the point detector is evaluated in terms of the evaluation mechanism published in the competition.
The trained models are tested on Synth2020 validation, UB PMC2020 validation and testset.

We compare our method with traditional image processing method, such as connected component analysis and detection-based method.
The detection model is based on Faster R-CNN.
To train the Faster R-CNN model, we expand the point $(x, y)$ into a rectangle $(x-r, y-r, x+r, y+r)$ whose data format is $(left, top, right, bottom)$.
We also implement another segmentation-based method Pose ResNet~\cite{xiao2018simple}, which is initially proposed for pose keypoint detection.
The Pose ResNet model adopts the structure of down-sampling and then up-sampling, without considering the feature fusion of different depths.

As listed in Table \ref{table:point detector},
the proposed method, which is simple and effective, outperforms other methods on three testsets.
As shown in Fig.~\ref{fig7},
on the Synth2020 validation set,
there are many cases where scatter points are connected and form a larger connected component.
On the UB PMC2020 testset, there are many noises in the plot area such as text elements.
Traditional image processing method can not distinguish close points which form a large component.
The detection-based method fails when the number of points is large or adjacent points are connected.
Compared to Pose ResNet, the feature fusion method helps to distinguish adjacent points, as shown in Fig.~\ref{fig7}(d).
The proposed method can effectively deal with these situations and locate adjacent points accurately.

\subsubsection{Robustness of feature on Data conversion}
We choose line type data to evaluate the performance of data conversion.
The performance of data conversion depends on the legend matching phase and value calculation phase.
The performance in the value calculation phase depends on whether OCR engine can recognize tick point value correctly.
Ignored the errors caused by the OCR engine, we discuss the robustness of extracted features in legend matching phase from the trained network.
As listed in Table \ref{table:data conversion},
we compare the performance when the legend matching phase is performed on the groundtruth and prediction result.
For short notation, 
here s1, s2, s3 represent average name score, average data series score, and average score, respectively, which is claimed in the evaluation script.

When using groundtruth as input,
the position of the elements is quite accurate.
Features extracted from trained network are comparable with features from the concatenation of RGB and HSV features.
The performance can be further improved by considering the cascading of features.
When prediction detection results are used, the position of elements may not be tight enough, which will bring in noise while extracting features.
Experiments show that features from our proposed method are more robust than image-based features.

\begin{table}[h]
    \footnotesize
    \centering
    \caption{Evaluation results of data conversion on line type data}
    \label{table:data conversion}

    \begin{tabular}[!hbt]{|c|c|c|c|c|c|c|}
        \hline
        \multirow{2}*{Features}   & \multicolumn{3}{|c|}{Groundtruth} & \multicolumn{3}{|c|}{Prediction} \\       
        \cline{2-7}
        & \ s1 \ & \ s2 \ & \ s3 \ & \ s1 \ & \ s2 \ & \ s3 \ \\
        \hline
        Baseline(RGB)    & \ 83.40 \ &\ 75.32 \ &\ 77.34 \ &\ 83.50 \ &\ 55.61 \ &\ 69.68 \ \\
        \hline
        RGB+HSV     & 83.48 & 77.58 & 79.06 & 83.50 & 52.52 & 68.16 \\
        \hline
        RGB+HSV+CNN & \textbf{83.48} & \textbf{78.31} & \textbf{79.60} & 83.06 & 53.81 & 68.83 \\
        \hline
        CNN & 82.71 & 77.78 & 79.02 & \textbf{83.53} & \textbf{67.19} & \textbf{75.36} \\
        \hline
    \end{tabular}
\end{table}

\begin{table}[h]
    \footnotesize
    \centering
    \caption{Evaluation results of proposed system}
    \label{table:system}
    \begin{threeparttable}
    \begin{tabular}[!hbt]{|c|c|c|c|c|c|c|c|c|}
        \hline
        \multirow{2}*{Chart types}   & \multicolumn{4}{|c|}{Synth2020 validation set} & \multicolumn{4}{|c|}{UB PMC2020 testset} \\       
        \cline{2-9}
        & \ s0 \ & \ s1 \ & \ s2 \ & \ s3 \ & \ s0 \ & \ s1 \ & \ s2 \ & \ s3 \ \\
        \hline
        Bar     &\ 93.36 \ &\ 99.82 \ &\ 99.11 \ &\ 99.29 \ &\ 91.75 \ &\ 96.96 \ &\ 94.20 \ &\ 94.89 \ \\
        \hline
        Scatter & 87.19  & 100.00 & 82.97 & 87.23 & 88.58 & 86.61 & 65.20 & 70.55 \\
        \hline
        Boxplot & 100.00 & 99.83 & 98.37 & 98.73 & 98.62 & 92.57 & 81.62 & 84.36 \\
        \hline
        Line    & 99.29  & 99.09 & 98.81 & 98.88 & 84.03 & 83.24 & 67.01 & 71.06 \\
        \hline
        \textbf{Average (Rank1*)} & - & - & - & - & 88.23 & \textbf{90.42} & 76.73 & 80.15 \\
        \hline
        \textbf{Average (Rank2*)} & - & - & - & - & 87.00 & 78.54 & 55.40 & 61.18 \\
        \hline
        \textbf{Average (proposed)} & \textbf{94.96} & \textbf{99.69} & \textbf{94.82} & \textbf{96.03} & \textbf{90.75} & 89.85 & \textbf{77.00} & \textbf{80.22} \\
        \hline
    \end{tabular}

    \begin{tablenotes}
        \footnotesize
        \item[*] from \url{https://chartinfo.github.io/leaderboards\_2020.html}
      \end{tablenotes}
    \end{threeparttable}
\end{table}

\subsubsection{Evaluation result of proposed system}
As listed in Table \ref{table:system}, we provide our proposed system performance on ICPR2020 Competition which can serve as a baseline and facilitate further research.
For short notation, here s0, s1, s2, and s3 represent visual element detection score, average name score, average data series score and average score, respectively.
In this work, no additional data or model ensemble strategy is adopted.
It is shown that our system outperforms the Rank1 and Rank2 result of the competition on UB PMC2020 testset, which demonstrate the effectiveness of the proposed system.

\section{Conclusion}
In this work, we discuss the data extraction stage in a data mining system.
For building a robust box detector, we compare different object detection methods and find a suitable method to solve the special issues that characterize chart data.
Models with multiple regression heads and FPN structure achieve impressive performance.
For building a robust point detector, compared with image processing-based methods and detection-based methods,
the proposed segmentation-based method can avoid hard heuristic assumptions and distinguish close points well.
For data conversion,
we propose a network to measure feature similarities which is more robust compared with image-based features.
In the experiments, we conduct experiments in each stage of data extraction.
We find the key factors that improve the performance of each stage.
The overall performance on a public dataset demonstrates the effectiveness of the proposed system.
Because an increasing number of charts have appeared in recent years,
we believe the field of automatic extraction from chart data will develop quickly.
We expect this work to provide useful insights and provide a baseline for comparison.

\section{Acknowledgement}
This research is supported in part by NSFC (Grant No.: 61936003, 61771199),  GD-NSF (no.2017A030312006).

\bibliographystyle{splncs04}
\bibliography{ref}

\end{document}